\documentclass[10pt,sigconf,letterpaper,nonacm]{acmart}
\usepackage{algorithmic}
\usepackage{graphicx}
\usepackage{textcomp}
\usepackage{enumitem}
\usepackage{multirow}
\usepackage{lscape}
\usepackage{subcaption}
\usepackage{makecell}
\usepackage{tabularray}
\newcommand\blfootnote[1]{%
  \begingroup
  \renewcommand\thefootnote{}\footnote{#1}%
  \addtocounter{footnote}{-1}%
  \endgroup
}

\expandafter\def\expandafter\normalsize\expandafter{%
    \normalsize%
    \setlength\abovedisplayskip{2pt}%
    \setlength\belowdisplayskip{8pt}%
    \setlength\abovedisplayshortskip{-8pt}%
    \setlength\belowdisplayshortskip{2pt}%
}
\usepackage[compact]{titlesec}
\titlespacing{\section}{0pt}{2ex}{1ex}
\titlespacing{\subsection}{0pt}{1ex}{0.5ex}
\titlespacing{\subsubsection}{0pt}{0.5ex}{0ex}
\AtBeginDocument{%
  }

\usepackage{xspace}
\newcommand{\PPTGNN}{\texttt{PPT-GNN}\xspace}

\begin{document}

%
\title{PPT-GNN: A Practical Pre-Trained Spatio-Temporal Graph Neural Network for Network Security}

\author{Louis Van Langendonck}
\email{louis.van.langendock@estudiantat.upc.edu}
\affiliation{%
  \institution{Universitat Polit{\'e}cnica de Catalunya}
  \city{Barcelona}
  \country{Spain}
}

\author{Ismael Castell-Uroz}
\email{ismael.castell@upc.edu}
\affiliation{%
  \institution{Universitat Polit{\'e}cnica de Catalunya}
  \city{Barcelona}
  \country{Spain}
}

\author{Pere Barlet-Ros}
\email{pere.barlet@upc.edu}
\affiliation{%
  \institution{Universitat Polit{\'e}cnica de Catalunya}
  \city{Barcelona}
  \country{Spain}
}

\renewcommand{\shortauthors}{Van Langendonck et al.}

\begin{abstract}

Recent works have demonstrated the potential of Graph Neural Networks (GNN) for network intrusion detection. Despite their advantages, a significant gap persists between real-world scenarios, where detection speed is critical, and existing proposals, which operate on large graphs representing several hours of traffic. This gap results in unrealistic operational conditions and impractical detection delays. Moreover, existing models do not generalize well across different networks, hampering their deployment in production environments. To address these issues, we introduce \PPTGNN, a practical spatio-temporal GNN for intrusion detection. \PPTGNN enables near real-time predictions, while better capturing the spatio-temporal dynamics of network attacks. \PPTGNN employs self-supervised pre-training for improved performance and reduced dependency on labeled data. We evaluate \PPTGNN on three public datasets and show that it significantly outperforms state-of-the-art models, such as E-ResGAT and E-GraphSAGE, with an average accuracy improvement of 10.38\%. 
Finally, we show that a pre-trained \PPTGNN can easily be fine-tuned to unseen networks with minimal labeled examples. This highlights the potential of \PPTGNN as a general, large-scale pre-trained model that can effectively operate in diverse network environments.


\end{abstract}

\ccsdesc[500]{Security and privacy~Network security}
\ccsdesc[500]{Computing methodologies~Artificial intelligence}
\ccsdesc[500]{Computing methodologies~Model development and analysis}
\ccsdesc[300]{Networks~Network monitoring}

\keywords{Temporal Graph Neural Network, Network Intrusion Detection, Few-shot Learning}


\maketitle

\section{Introduction}
The rapid evolution of computer network systems and the critical infrastructure they support highlights the importance of robust Network Intrusion Detection Systems (NIDS). Traditional machine learning-based solutions work well on flat data, but fail to capture structural dependencies between network traffic flows \cite{gnnsurvey}. This results in less efficient detection and increased vulnerability to adversarial attacks. To address these challenges, several recent works proposed the use of Graph Representation Learning (GRL) and Graph Neural Networks (GNNs) \cite{resgat, tsids, anomale, egraphsage}. By depicting network traffic as a graph, both flow-level features and structural properties are used in decision making, leading to more accurate and robust intrusion detection \cite{gnnsurvey}. 

However, despite their good performance, a large gap can be observed between theoretical evaluation on standard datasets and the practical challenges involved in real-world deployment \cite{outsidetheclosedworld, feasability, alwayspretraining}. We argue that there are two main reasons for this discrepancy. Firstly, these solutions require extensive training from scratch for each network environment, hence requiring accurately labeled data specific to the target network. This data is difficult to acquire, as it usually requires network instrumentation and manual inspection. Additionally, continuously evolving attacks need repeated data acquisition and model training, which further complicates deployment in operational networks. Secondly, training and validation splits are usually obtained by randomly sampling a large static graph representing the traffic of several hours or days \cite{feasability, resgat, egraphsage}. 
This is problematic because detection engines need to observe a substantial portion of the incoming traffic, causing detection delays and hindering online detection. Random sampling may also result in unrepresentative traffic, breaking structural and dynamic patterns, and causing data leakage from train to test sets as flows from the same attack can be present in different splits.

In this work we propose \PPTGNN, a new model for intrusion detection that includes two key elements to address these limitations: $(i)$ a novel graph-learning method that is trained and evaluated on temporal sliding window snapshots of up to a only few seconds, and $(ii)$ a self-supervised pre-training step, which does not require labeled data. The former captures the dynamics and temporal dependencies of network traffic flows by aggregating inter- and intra-snapshot temporal information. The latter is used to address the data acquisition and network heterogeneity issues in real deployments. This is achieved by pre-training \PPTGNN on a link-prediction task on a reconciled, unlabeled dataset. 
We explore two pre-training strategies: one where unlabeled data from the target network itself is utilized to pre-train the model (in-context pre-training), and one where unlabeled data samples from the target network are exclusively used for fine-tuning an already pre-trained model (out-of-context pre-training).
To evaluate our models under realistic conditions, we build flow-based datasets using the raw packet-level traces from the publicly available UNSW-NB15 \cite{unswnb15}, ToN-IoT \cite{toniot} and BoT-IoT \cite{botiot} datasets. We restricted the set of flow features in our datasets to those that can be directly derived from standard NetFlow v9 records. In contrast, the publicly available flow-based versions of these datasets \cite{nfstandard} (used in previous works) contain information that would not be available in real flow-level deployments (e.g., packet size distribution, DNS information or TCP statistics). We argue that using unlabeled NetFlow-compatible data eases deployment, and allows for continuous self-supervised training, accommodating for the changing nature of network traffic. Finally, using few-shot learning, we show that pre-training on out-of-context network data, followed by fine-tuning on a specific network setting, significantly reduces both training times and the required amount of labeled data.

Specifically, this paper makes the following contributions:
\begin{itemize}
    \item We propose \PPTGNN, a practical spatio-temporal GNN model for intrusion detection that can operate under realistic operational conditions.
    \item We propose a self-supervised pre-training process that improves detection performance, decreases training times and reduces dependency on labeled data.
    \item We show that a pre-trained \PPTGNN with out-of-context data can easily be fine-tuned to new network scenarios, significantly reducing deployment costs.
    \item We show that \PPTGNN improves the multi-class Macro F1 score of \textit{E-ResGAT} \cite{resgat} by 10.83\% and \textit{E-GraphSAGE} \cite{egraphsage} by 9.93\%, despite they operate under unrealistic (less challenging) conditions.
\end{itemize}

The rest of the paper is organized as follows: Section \ref{related_work} reviews the related work. Section \ref{methodology} describes the fundamentals of our new model proposal. Section \ref{evaluation} presents the results of our evaluation and Section \ref{conclusions} concludes the paper.

\section{Related work}
\label{related_work}
The research community already proposed several NIDS solutions based on traditional Machine Learning algorithms, such as Random Forests (RF), K-nearest neighbors (KNN) or Support Vector Machines (SVM). Consequently, there are numerous surveys that cover the related work in this area (e.g., \cite{khraisat_survey_2019, resende_survey_2018, ghorbani_network_2009, garcia-teodoro_anomaly-based_2009}). In this section we will focus only on the most recent proposals using Graph Neural Networks (GNN).

The most representative GNN frameworks for network intrusion detection are arguably E-GraphSAGE, proposed by Lo et al. \cite{egraphsage}, and E-ResGAT, put forth by Chang et al. \cite{resgat}. The E-GraphSAGE method builds simple network graphs, where nodes denote IP-port combinations and the edges represent the flows connecting them. E-GraphSAGE extends the GraphSAGE algorithm to handle edge classification using node feature concatenation. On several publicly available datasets, an average increase of $2\%$ in weighted F1 over previous state-of-the art is obtained. The E-ResGAT method converts the network representation to a line graph, in which flows are represented as intermediary nodes. They apply an attention mechanism with residual connections to aggregate the two-hop neighbourhood and update central node representations.  
Nevertheless, both of them suffer from the same problem, the necessity of complete graphs spanning hours or even days of traffic to train their models, a fact that does not allow them to perform online detection.


Regarding the practicality of GNN-based NIDS systems, Gu et al.~\cite{alwayspretraining} propose an encoder-decoder architecture to learn dense representations of categorical features, which can reduce the dependency on labeled data.
However, they still require full graphs and in-context data to train the auto-encoder.
On the other hand, Venturi et al.~\cite{feasability} were the first to raise concerns surrounding the detection delay problem in practical implementation of GNNs. They evaluated a binary E-GraphSAGE model on different smaller time frames, ranging from seconds to hours. They demonstrate that detection performance drops significantly for several attack types, rendering the GNN impractical for online detection. This stands as one of our main motivations to propose a new method that can actually operate in more realistic conditions.

\section{Methodology}
\label{methodology}

In this section we introduce a new approach for intrusion detection that addresses several challenges involved in practical deployment of GNN-based NIDS. 
Here, we refer to the \textit{network intrusion detection task} as a traffic classification problem where, for each flow seen by a NIDS, the system needs to determine whether the flow corresponds to an attack or to legitimate traffic. In case the flow is considered malicious, the NIDS should also determine the attack type.

Our proposal for intrusion detection (\PPTGNN) is based on a combination of three key elements: $(i)$ a novel graph structure that represents both the spatial and temporal dimensions of network traffic, $(ii)$ a GNN that exploits this new graph representation, and $(iii)$ a self-supervised pre-training strategy that allows for efficient deployment, using large-scale training on `cheap', unlabeled network data, while minimizing the need of in-context, attack-labeled data. 
The rest of this section describes each component in detail.

\subsection{Flow Graph Construction} \label{graph-const}
Flow-level traffic can be represented as a heterogeneous graph. For the spatial component of the graph, we use the source and destination IP addresses of each flow to represent the source and destination nodes, respectively. We opt for the line-graph representation for flows, which means that instead of representing the flow as an edge with edge attributes, we initialize the flow as a node with its flow features as attributes, connected with featureless edges to its corresponding IP nodes, turning the problem of intrusion detection into one of node classification. 
The edges between flow nodes and IP nodes are bidirectional and each considered separately in the heterogeneous graph. This makes up the four spatial connections in the graph building process.

\begin{figure*}[t]
    \centering
    \includegraphics[width=00.75\textwidth]{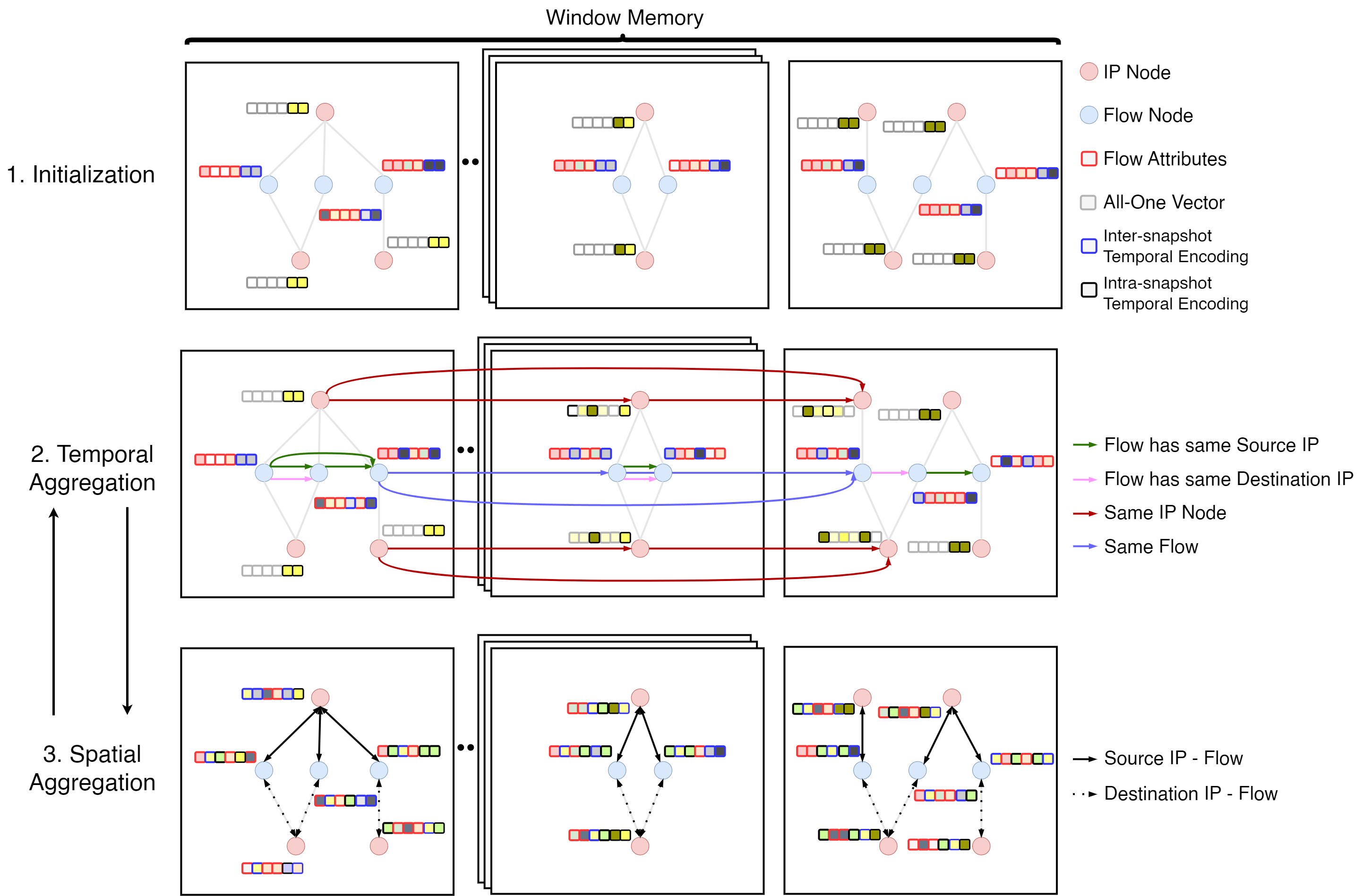}
    \caption{\textmd{Schematic overview of the proposed spatio-temporal \PPTGNN architecture.}}
    \Description[<short description>]{<long description>}
    \label{fig:diagram_v2}
\end{figure*}

However, using the aforementioned method to represent the entire flow graph (which may span several hours or days) and using it for training, validation and testing, introduces a range of practical problems. On the one hand, it implies the entire graph is available at inference time. This can clearly not be assumed in a practical scenario, where the flow graph would be gradually built as network flows arrive and leave the network, and intrusions should be detected as soon as they occur. This introduces the question of when the model should make the predictions. Executing the model online may lead to poor performance due to the difference in scale between the small flow graphs seen in practice and those in training, which GNNs typically struggle to handle \cite{scale}. Postponing prediction, however, would result in detection delays, rendering the NIDS useless. On the other hand, the dynamic, temporal behaviour of network traffic can provide useful information for the detection task, but it is not considered in the static graph approach. Additionally, the random train, validation and test splitting used in previous works breaks some of these temporal dynamics and relationships between nodes and flows in the graph.

Therefore, we propose to introduce a sliding time window approach with temporal encoding, and both \textit{intra-window} and \textit{inter-window} temporal edges. 
Figure~\ref{fig:diagram_v2} presents a high-level overview of this framework.
With a chosen window size of up to a few seconds, we slide over the dataset, building the flow graph for all flows and corresponding IPs that start or end within this window. Note that longer duration flows appear in multiple time windows. Splitting on time instead of on the number of flows also implies a high variance in the amount of nodes seen per graph in training, hence resulting in a more diverse, realistic dataset. 

Now the heterogeneous graph is extended to contain temporal edges. In terms of inter-window connections, edges are created between IPs and flows that reappear in different windows (red and blue arrows in Figure~\ref{fig:diagram_v2}, respectively). However, to limit the amount of these types of connections spanning across the entire dataset, we allow each window to only access a fixed amount of previous windows, called the \textit{window memory}. Now, within this window memory, each reappearing IP and flow is connected to all its recurrences in the following windows. Finally, concerning intra-window temporal edges, two types are introduced. First, all flow nodes within a time window connected to the same source IP are inter-connected, in their temporal order (green arrows). The same applies for all flow nodes in a window connected to the same destination IP (pink arrows). Again, to limit the amount of connections made, we impose a maximum amount of previous temporal connection for each flow, called the \textit{flow memory}. 

\subsection{Spatio-temporal Model} \label{model}

Now we propose a Graph Neural Network architecture that exploits the temporally enriched flow graph topology. We adopt a multi-step, hierarchical approach inspired by the HTGNN framework \cite{htgnn}. To capture spatial, temporal, intra-window, and inter-window dependencies, we implement multiple heterogeneous spatio-temporal layers. Each layer performs consecutive temporal and spatial node updating steps.

First, we initialize the graph $\mathcal{G}(\mathcal{V}, \mathcal{S}, \mathcal{T})$ as a collection of interconnected heterogeneous subgraphs $\mathcal{G}_t'(\mathcal{V}, \mathcal{E}, \mathcal{F})$ for timestamps $t \in (T - t_{window\_size}, \ldots, T)$ and edge set $\mathcal{F}'$, where $\mathcal{E}$ is the spatial edge set, $\mathcal{F}$ is the temporal intra-window edge set and $\mathcal{F}'$ contains inter-window temporal edges connecting the different subgraphs in memory. Together, these form the full graph $\mathcal{G}(\mathcal{V}, \mathcal{S}, \mathcal{T})$, with $\mathcal{S}$ representing all spatial edges and $\mathcal{T}$ representing all temporal edges (both intra-window and inter-window). 

Each node $v$ has an associated intial hidden state $h_{v, 0}$, which is initialized based on the node type: Flow nodes use feature vectors from the dataset, while IP nodes are initialized with dummy one-vectors (note that IP addresses are not stored either in IP nodes or flow nodes to avoid overfitting). To enrich these representations, we concatenate temporal information using cyclical encoding vectors. Flow nodes receive a 30-dimensional encoding that captures the order of flows within a window, and IP nodes receive an encoding that indicates the position of the window within the memory (Step 1 in Figure \ref{fig:diagram_v2}).


In each layer of our framework, we first perform a temporal node updating step (Step 2 in Figure \ref{fig:diagram_v2}). For each node $v$ involved in temporal edges in $\mathcal{T}$, we update its hidden state as follows. For each temporal edge type $e$ in $\mathcal{T}$, we compute intermediate hidden states as:

\[
h_{v, e}^{\text{temp}} = W_1^{(e)} h_v + W_2^{(e)} \oplus_1 \left( h_u \mid u \in N_v^e \right),
\]

where \(\oplus_1\) represents the first aggregation method (e.g., mean, sum, or max), \(W_1^{(e)}\) and \(W_2^{(e)}\) are edge type-specific weight matrices, and \(N_v^e\) denotes the neighbors of node \(v\) for edge type \(e\). These intermediate states \(h_{v, e}^{\text{temp}}\) are collected into a list \(\text{temp\_agg\_list}_v\). The final node representation is obtained by applying the second aggregation method \(\oplus_2\) over this list and then applying a non-linearity:

\[
h_v^{k_{temp}} \leftarrow \sigma \left( \oplus_2 \left( \text{temp\_agg\_list}_v \right) \right).
\]

Next, we perform a spatial node updating step (Step 3 in Figure \ref{fig:diagram_v2}). Using the hidden states obtained from the temporal step, we update each node $v$ involved in spatial edges in $\mathcal{S}$. For each spatial edge type $e$ in $\mathcal{S}$, we compute intermediate hidden states as follows:

\[
h_{v, e}^{\text{spatial}} = W_1^{(e)} h_v^{k_{temp}} + W_2^{(e)} \oplus_1 \left( h_u^{k_{temp}} \mid u \in N_v^e \right),
\]

These intermediate states \(h_{v, e}^{\text{spatial}}\) are collected into a list \(\text{spatial\_agg\_list}_v\). The final node representation is obtained by applying the second aggregation method \(\oplus_2\) over this list and then applying a non-linearity:

\[
h_v^k \leftarrow \sigma \left( \oplus_2 \left( \text{spatial\_agg\_list}_v \right) \right).
\]

By stacking such layers of consecutive temporal and spatial node updating steps, spatio-temporal information is effectively propagated through the network. This results in a highly expressive spatio-temporal graph learning framework. The final node representations are used as input to a multi-layer perceptron for multi-class classification to predict flow labels. Although the order of temporal and spatial aggregation can be interchanged, prioritizing temporal aggregation leverages temporal encoding immediately after initialization, enhancing the preservation of temporal information before spatial message-passing.

\subsection{Self-supervised Pre-training}

The final key step of \PPTGNN is a self-supervised pre-training process that, using a large amount of unlabeled data, can find feasible transformations on our graph data to learn from. This allows for using less expensive labeled data to achieve similar performance. 

Several graph-based self-supervised pre-training strategies have been proposed \cite{gnnpretrainingstrategies}, ranging from context prediction and attribute prediction to graph-level property prediction. We claim that negative edge sampling and link prediction are effective for learning spatio-temporal dynamics. First, for a given temporal heterogeneous graph, negative edge sampling is applied. This implies adding negative edges, of the different spatial and temporal types, that are not initially present in the graph. The remaining edges are considered the positive examples. We thus define the self-supervised link-prediction task as a binary classifier that predicts, for every edge in the graph, if that edge is positive or negative. We employ the GNN architecture from Section \ref{model}, replacing the multi-class classifier for a binary one, and use the final weights of the pre-trained model as base for the intrusion detection task. This pre-training task suits the problem by forcing the model to understand the roles of different edge types, capturing fundamental network dynamics. Now the question arises: which unlabeled data to use for pre-training? A first choice is \textit{in-context pre-training}, which implies using the data available of the target network. This can yield favorable representations specific to that target, though it limits generality and increases deployment cost and complexity. Therefore, a preferable approach is using \textit{out-of-context pre-training}, where a large amount of data from other networks is used, providing the pre-trained model a more general understanding of network traffic, that can later be fine-tuned to a specific network context. Both options are explored in Section \ref{eval:results}

\begin{table}
\caption{\textmd{Performance of the \PPTGNN model compared to baseline and state-of-the-art models. Best performing model for every dataset-metric combination is highlighted in bold.}}
\label{results_table}
\resizebox{\columnwidth}{!}{%
\begin{tabular}{llllll}
\hline
Dataset&
  Model&
  \multicolumn{2}{l}{\textbf{F1 Weighted}} &
  \multicolumn{2}{l}{\textbf{F1 Macro}} \\ &
   &
  Binary &
  Multi &
  Binary &
  Multi \\ \hline
NF-UNSW-NB15&
  MLP &
  \textbf{0.981} &
  0.943 &
  0.658 &
  0.223 \\
 &
  E-GraphSAGE&
  0.976 &
  0.963 &
  \textbf{0.913} &
  0.415 \\
 &
  E-ResGat &
  0.974 &
  0.960 &
  0.853 &
  0.363 \\
 &
  PPT-GNN&
  0.977 &
  \textbf{0.966} &
  0.806 &
  \textbf{0.482} \\ \hline
NF-ToN-IoT&
  MLP 
&
  0.921 &
  0.943 &
  0.904 &
  0.796 \\
 &
  E-GraphSAGE&
  0.910 &
  \textbf{0.964}&
  0.920 &
  0.793 \\
 &
  E-ResGat 
&
  0.979 &
  0.960 &
  0.920 &
  0.790 \\
 &
  PPT-GNN&
  \textbf{0.983}&
  \textbf{0.964}&
  \textbf{0.967}&
  \textbf{0.935}\\ \hline
NF-BoT-IoT&
  MLP 
&
  0.920 &
  0.478 &
  0.902 &
  0.388 \\
 &
  E-GraphSAGE&
  0.790 &
  0.589 &
  0.737 &
  0.524 \\
 &
  E-ResGat 
&
  0.895 &
  \textbf{0.641} &
  0.862 &
  0.552 \\
 &
  PPT-GNN&
  \textbf{0.928} &
  0.456 &
  \textbf{0.908} &
  \textbf{0.613} \\ \hline
\end{tabular}%
}
\end{table}

\section{Evaluation}
\label{evaluation}
\subsection{Dataset}
\label{eval:dataset}
Network intrusion datasets often suffer from a lack of quality labeled data, standardization, and realism in feature sets \cite{outsidetheclosedworld}. To address these issues, we create standardized datasets from raw network data captures of popular datasets: UNSW-NB15 \cite{unswnb15}, ToN-IoT \cite{toniot}, and BoT-IoT \cite{botiot}. Although there already exist public ``NetFlow'' versions of these datasets \cite{nfstandard}, they include many features impossible to obtain exclusively from the information available in most flow-level monitoring systems (e.g., per-flow packet-size distribution, DNS information or TCP statistics). In contrast, we value the practicality of the system and its viability to be deployed in real environments using NetFlow or IPFIX standard. We carefully select features directly derivable from NetFlow v9 records to ensure practical applicability. Moreover, this standardization enables cross-dataset pre-training, addressing labeled data scarcity. 

\subsection{Performance results}
\label{eval:results}

Our first experiment compares \PPTGNN to both a baseline and the state-of-the-art (SOTA). As baseline, we opt for a Multi-layer Perceptron (MLP), given it is practically equivalent to a GNN without topology. In terms of SOTA, we implement both the E-GraphSAGE and E-ResGAT methods described in Section~\ref{related_work}. 
We train and hyperoptimize each of the aforementioned methods on the same \textit{train} and \textit{validation} split of the datasets described in Section \ref{eval:dataset}. For the \PPTGNN model, we consider window time sizes ranging from 0.5 to 20 seconds, keeping from 1 to 10 windows in memory and fixing the flow memory to maximum 20 nodes. For more information on the hyperparameter optimization of \PPTGNN, we refer to Appendix \ref{appendix:hyper}. For each algorithm and dataset combination, we select the best performing model based on predefined criteria and evaluate it on the same test set. Evaluation uses four metrics: multi-class and binary weighted F1, and macro F1. The multiclass macro F1 score, which reflects unweighted performance over all classes, poses a particular challenge due to the significant presence of minority classes in the datasets. We optimize parameters based on this metric, as it best reflects the model’s overall understanding.  An overview of the results is depicted in Table \ref{results_table}. 
We can observe a significant improvement in the multiclass macro F1 score compared to the respective second-best performing models on each dataset: by 6.7\%, 13.9\%, and 5.9\% on the NF-UNSW-NB15, NF-ToN-IoT, and NF-BoT-IoT datasets, respectively. This improvement is achieved using a limited feature subset and few-second graphs, unlike other GNN models that use full graphs and unavailable features in standard flow-level data. We attribute this improvement to $(i)$ the temporal understanding of our proposed network architecture and $(ii)$ the pre-training step that reduces overfitting.  
For a more comprehensive understanding of these outcomes, we refer readers to the confusion matrix comparisons depicted in Appendix \ref{appendix:confusion}, where improvements among different attack classes are clearly visible.

\begin{figure}[t]
    \includegraphics[width=\columnwidth]{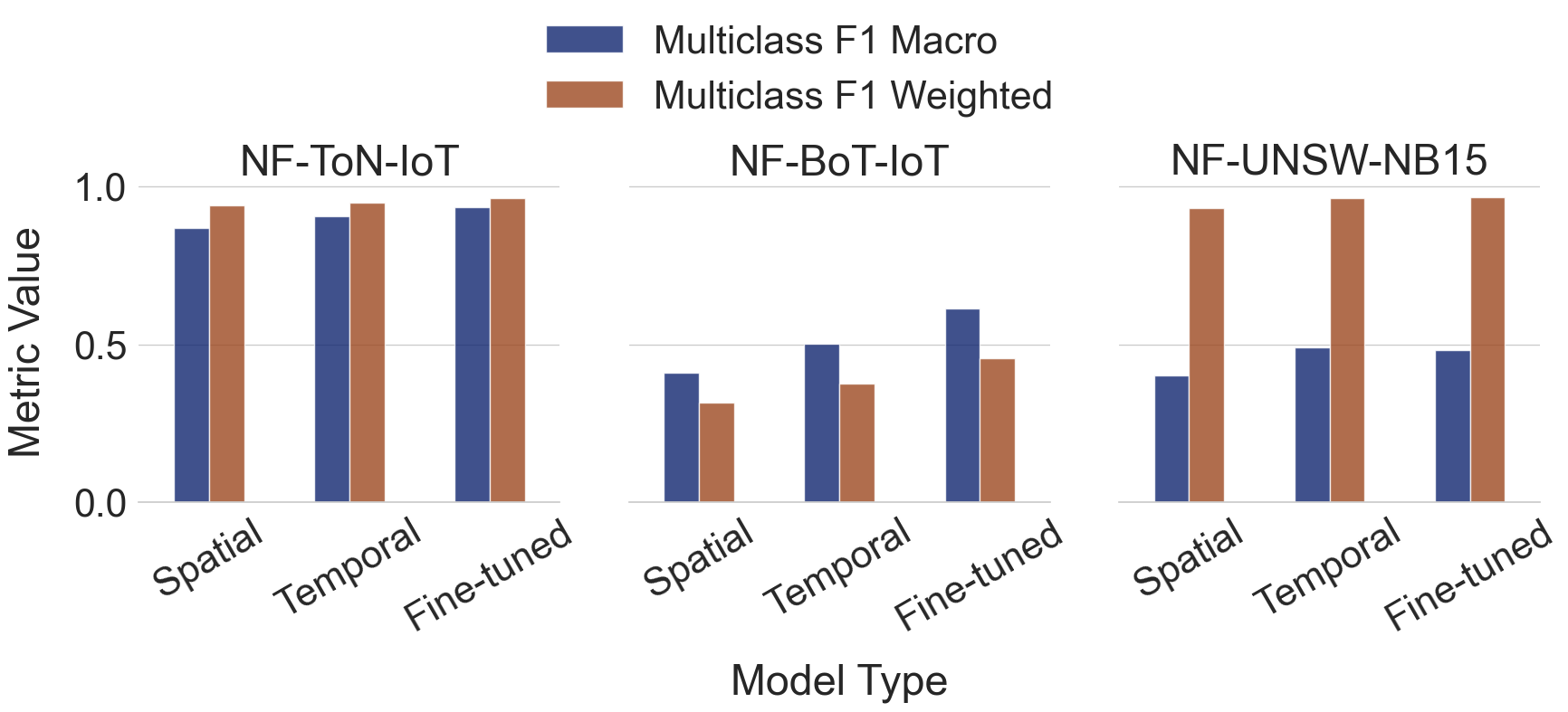}
    \caption{\textmd{Performance comparison showing the impact of iteratively adding key components of the \PPTGNN framework for each dataset, and tracking changes in multiclass metrics.}}
    \Description[<short description>]{<long description>}
    \label{fig:model_breakdown_improved}
\end{figure}

\subsection{Differential analysis}

To gain a more thorough understanding of the \PPTGNN model internals, we perform a differential analysis on its main components. We first train and optimize a purely spatial model without temporal edges on individual time windows. Then we add temporal edges and optimize our model again. Finally, the same hyperparameter setup is used for in-context pre-training and subsequent fine-tuning. The corresponding results are displayed in Figure \ref{fig:model_breakdown_improved}. Though a general upwards trend proportional to model complexity is observed across all datasets, we note that the NF-BoT-IoT and NF-UNSW-NB15 datasets benefit more from the use of temporal dynamics and pre-training than the NF-ToN-IoT one.  

\begin{figure}[t]
    \includegraphics[width=\columnwidth]{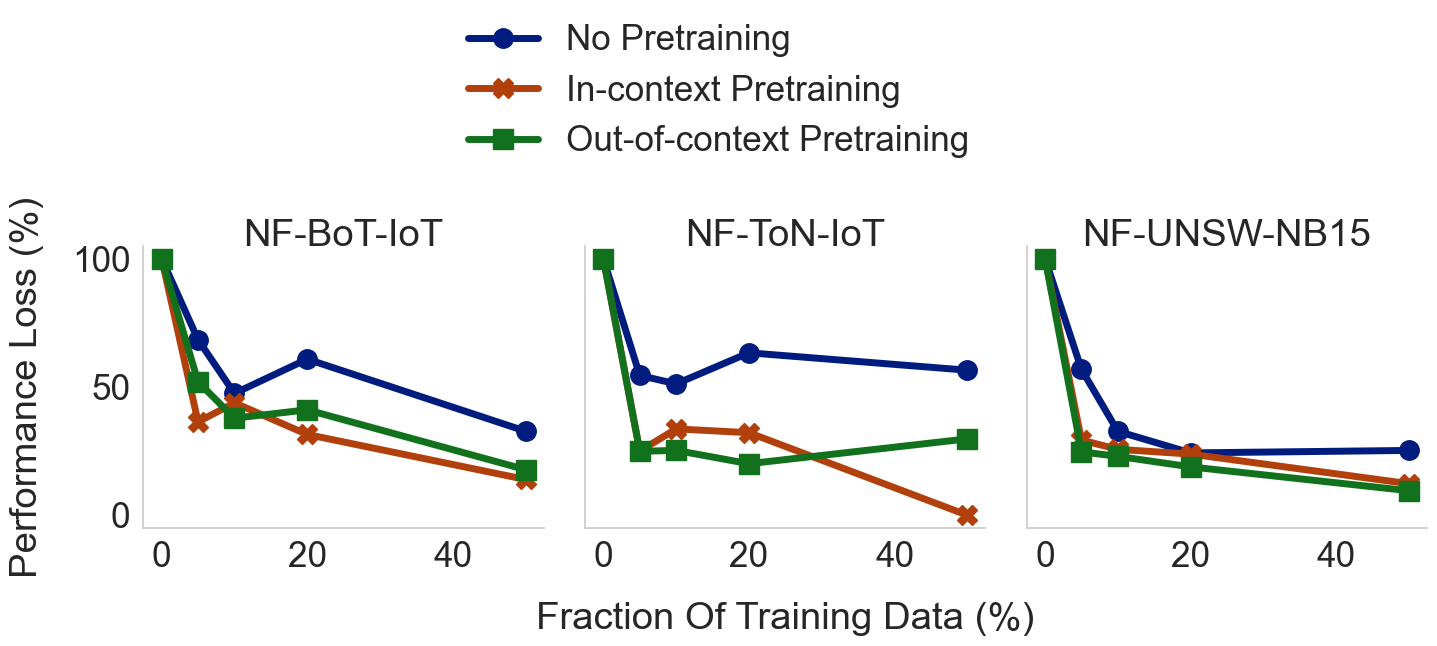}
    \caption{\textmd{Results of few-shot learning experiments: For each pre-training strategy, we train on small training data fractions and track the percentual performance loss in multiclass macro-F1 score relative to the best overall performing \PPTGNN model for each dataset.}}
    \Description[<short description>]{<long description>}
    \label{fig:k_shot_learning}
\end{figure}

\subsection{Pre-training and Fine-tuning results}

For the pre-training experiments, we use the \PPTGNN hyperparameter set that, when trained from scratch, yields on average the best results across all datasets. The resulting model setting considers a time window size of 5 seconds, a window memory of 5 and a flow memory of 20 nodes. To display the generalization prowess of this architecture, we set up a few-shot learning experiment where in a fine-tuning context, using a pre-trained GNN base, we limit both the amount of labeled training data and number of epochs. The experiment considers, for each dataset, three pre-training scenarios: $(i)$ In-context pre-training, using unlabeled data of the dataset in question, $(ii)$ out-of-context pre-training where data from the remaining two datasets was used and $(iii)$ no pre-training. For each scenario and dataset, we track the percentual performance loss in multiclass macro-F1 score with respect to the best overall performing \PPTGNN model for that dataset. A temporal undersampling strategy that maintains class-balance is used to obtain the fractions of the training data, ranging from 0\% to 50\% of the available labeled data. The amount of epochs is reduced from 200, used for training from scratch, to 50. The results of this experiment are displayed in Figure \ref{fig:k_shot_learning}.

The results clearly show that pre-training accommodates for more efficient learning. Using any of the pre-training strategies, a cross-data average 68\% of optimal performance is achieved training on just 5\% of available labeled data, compared to just about 40\% of performance without pre-training. More remarkably, out-of-context pre-training yields results comparable to those using in-context pre-training. This surprising result implies that during pre-training, despite each network setting having significantly different behavior, the \PPTGNN architecture captures context-agnostic dynamics that can generalize to unseen scenarios.
This suggests that large-scale pre-training on unlabeled network data could provide a general purpose, transferable framework, similar to that obtained by Large Language Models \cite{chatgpt}.

To quantify the benefits of using a pre-training step, we compare fine-tuning time against training from scratch. Table \ref{tab:time_results_table} presents out-of-context, fine-tuning training times, alongside training times from scratch using all labeled data. Averaging results across all datasets, we conclude fine-tuning can yield reductions in training time of up to nearly 95\%.

\begin{table}
\caption{\textmd{Comparison of training times: fine-tuning vs. training from scratch.}}
\label{tab:time_results_table}
\resizebox{\columnwidth}{!}{%
\begin{tabular}{llllll}
\hline
\begin{tabular}[c]{@{}l@{}}Training \\ Type\end{tabular} &
  \begin{tabular}[c]{@{}l@{}}Training \\ Fraction\end{tabular} &
  \multicolumn{3}{c}{\textbf{Training Duration (s)}} &
  \textbf{\begin{tabular}[c]{@{}l@{}}Proportional\\ Average\\ Duration (\%)\end{tabular}} \\
                                                                                   &      & NF-UNSW-NB15 & NF-ToN-IoT & NF-BoT-IoT & \multicolumn{1}{c}{} \\ \hline
\multirow{4}{*}{\begin{tabular}[c]{@{}l@{}}Fine-tuning\\ (50 epochs)\end{tabular}} & 0.05 & 103.29       & 88.02      & 54.17      & 5.54                 \\
                                                                                   & 0.1  & 121.77       & 103.64     & 57.08      & 6.28                 \\
                                                                                   & 0.2  & 167.79       & 138.73     & 71.05      & 8.31                 \\
                                                                                   & 0.5  & 250.94       & 273.89     & 137.77     & 14.72                \\ \hline
\begin{tabular}[c]{@{}l@{}}From Scratch\\ (200 epochs)\end{tabular}                & 1.0  & 1666.34      & 1923.60    & 927.39     & 100.00               \\ \hline
\end{tabular}%
}
\end{table}

\section{Conclusions}
\label{conclusions}

In this work, we proposed \PPTGNN, a graph-based Network Intrusion Detection System (NIDS) designed to bridge the notable gap between recent GNN-based proposals and their practical implementation in a real environment. To tackle this problem, we introduce two key elements, a new graph representation that incorporates both spatial and temporal relationships between flows by means of temporal sliding window snapshots, and a self-supervised pre-training step that allows us to train the model on unlabeled datasets for subsequent fine-tuning on a target network. Our results obtained on three different flow-based datasets show that \PPTGNN obtains in average more than a 10\% improvement in the Macro F1 metric with respect to E-GraphSAGE and E-ResGAT, despite \PPTGNN operates under more realistic and challenging conditions. 
Remarkably, our findings show that pre-training the \PPTGNN architecture on diverse network settings yields results comparable to those obtained with specialized pre-training. 
These results suggest that large-scale pre-training on unlabeled network data could provide a general framework that considerably lowers deployment barriers, reducing model training and maintenance costs, increasing model transferability, and decreasing the dependency on labeled data. Labeled data sets are scarce and difficult to obtain in this area, as they often require significant manual inspection and curation by cybersecurity experts.


\blfootnote{This paper is currently under review. All code will be published upon acceptance of the paper.}
\blfootnote{This work does not raise any ethical issues.}


\bibliographystyle{ACM-Reference-Format}
\bibliography{bib}

\newpage
\appendix
\onecolumn
\section{Hyperparameters}\label{appendix:hyper}

\begin{table}[h]
\centering
\caption{\textmd{Overview of hyperparameters used for model optimization and training.}}
\label{appendix:hypertable}
\resizebox{0.52\columnwidth}{!}{%
\begin{tblr}{
  column{even} = {c},
  column{3} = {c},
  cell{1}{2} = {c=3}{},
  hline{1,3,17} = {-}{},
}
                                     & \textbf{Value(s)}     &                           &                    \\
                                     & From Scratch          & Pre-training              & Fine-tuning        \\
\textbf{Window Size (s)}             & {[}20, 10, 5, 1, 0.5] & 5                         & 5                  \\
\textbf{Window Memory (No. Windows)} & {[}1, 3, 5]           & 5                         & 5                  \\
\textbf{Flow Memory (No. Flows)}     & 20                    & 20                        & 20                 \\
\textbf{GNN Layers}                  & {[}2,3]               & 2                         & 2                  \\
\textbf{GNN Hidden Size}             & 128                   & 128                       & 128                \\
\textbf{Classifier Layers}           & 2                     & 2                         & 2                  \\
\textbf{Classifier Hidden Size}      & 128                   & 128                       & 128                \\
\textbf{Learning Rate}               & 0.001                 & 0.0001                    & 0.01               \\
\textbf{Activation Function}         & Leaky ReLU            & Leaky ReLU                & Leaky ReLU         \\
\textbf{SAGEConv Aggregation}        & {[}sum, mean]         & mean                      & mean               \\
\textbf{HeteroConv Aggregation}      & sum                   & sum                       & sum                \\
\textbf{Loss Function}               & Cross-Entropy Loss    & Binary Cross-Entropy Loss & Cross-Entropy Loss \\
\textbf{Weighted Loss}               & {[}True, False]       & False                     & True               \\
\textbf{Optimizer}                   & Adam                  & Adam                      & Adam               
\end{tblr}}
\end{table}

\section{Confusion matrices}\label{appendix:confusion}

\begin{figure*}[!h]
    \centering
    \includegraphics[width=0.59\textwidth]{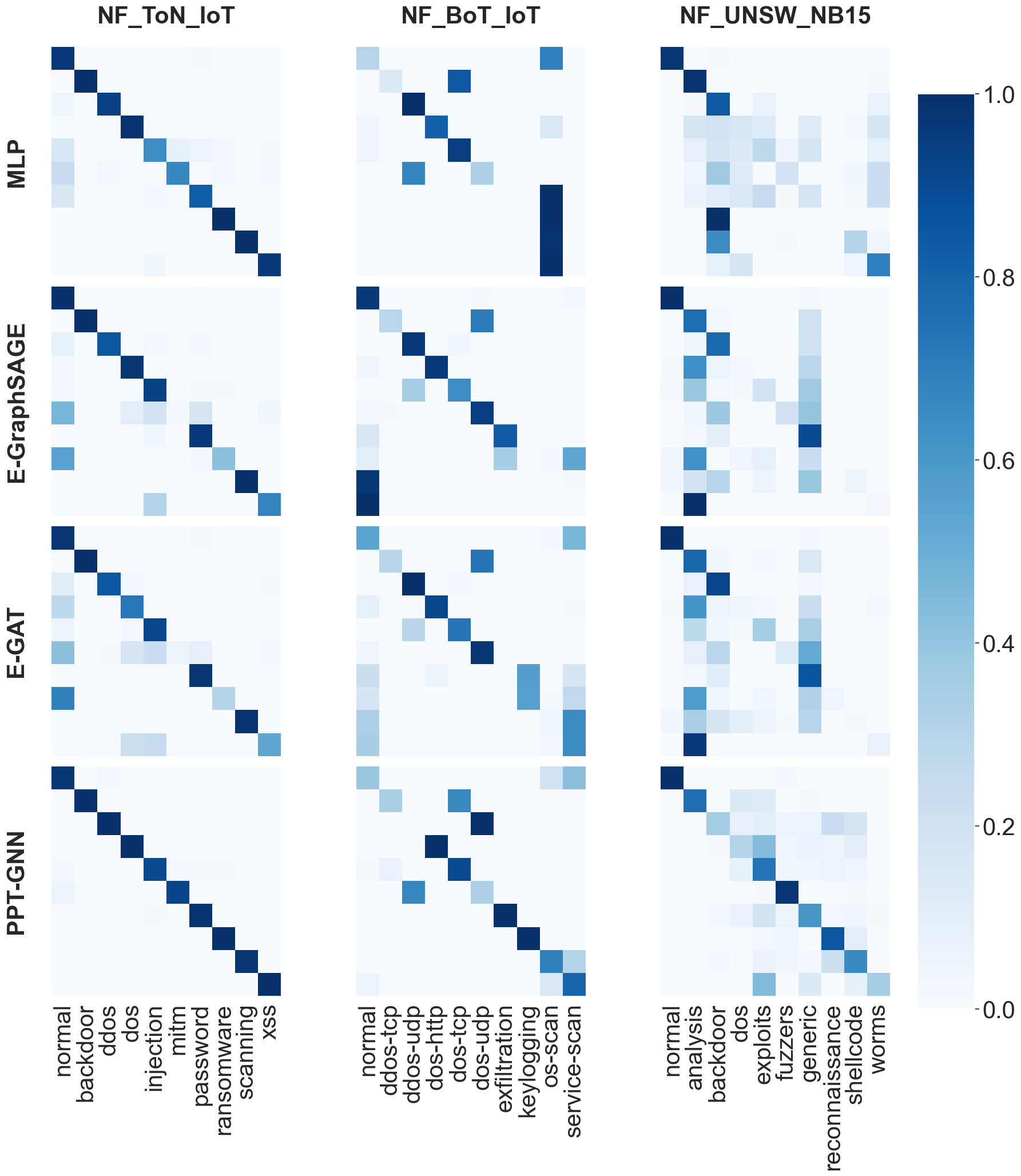}
    \caption{Confusion matrices. \textmd{Values are normalized on the model predictions. The diagonal of each matrix represents the match between the actual and the predicted attack. }}
    \label{fig:matrices}
\end{figure*}

\end{document}